\documentclass[letterpaper, 10 pt, conference]{ieeeconf}  

\IEEEoverridecommandlockouts                              

\usepackage{amsmath} 
\usepackage{graphicx}
\usepackage{color}
\usepackage{caption}
\usepackage{subcaption}
\usepackage{float}
\usepackage[absolute,overlay]{textpos}
\title{\LARGE \bf
Validation of two-wire power line UAV localization based on the magnetic field strength
}

\author{
Goran Vasiljevi{\'{c}}$^{1}$, Dean Martinovi{\'{c}}$^{1}$, Matko Bato{\v{s}}$^{1}$ and Stjepan Bogdan$^{1}$
\thanks{}
\thanks{
$^{1}$Goran Vasiljevi{\'{c}}, Dean Martinovi{\'{c}}, Matko Bato{\v{s}} and Stjepan Bogdan are with University of Zagreb
Faculty of Electrical Engineering and Computing, Laboratory for Robotics and Intelligent Control Systems (LARICS - https://larics.fer.hr/larics), Unska 3, Zagreb 10000, Croatia; 
        {\tt\small goran.vasiljevic@fer.hr}}
        }

\begin{document}

\maketitle
\thispagestyle{empty}
\pagestyle{empty}
\maxdeadcycles=20000
\begin{textblock*}{14.9cm}(3.2cm,0.75cm) %
	{\footnotesize © 2022 IEEE.  Personal use of this material is permitted.  Permission from IEEE must be obtained for all other uses, in any current or future media, including reprinting/republishing this material for advertising or promotional purposes, creating new collective works, for resale or redistribution to servers or lists, or reuse of any copyrighted component of this work in other works.}
\end{textblock*}

\begin{abstract}

In this paper we extend our previous work on UAV localization based on the magnetic field strength. The method is based on a magnetic flux density distribution in vicinity of two very long, thin and parallel transmission lines. An UAV is equipped with 4 magnetometers, positioned so that obtained measurements give unique solution to an optimization problem used to find relative position and orientation of the UAV with respect to conductors. Several sets of experiments, undertaken on a laboratory setup, confirmed validity of the method for both solutions - analytical and numerical optimization. Obtained results, compared with high precision motion capture system, are within range of standard RTK positioning.
\end{abstract}

\section{INTRODUCTION}

In our previous work \cite{martinovic_mathematical_2021} it is shown that unmanned aerial vehicles (UAVs) can precisely navigate through the magnetic field of two long parallel power lines using at least three magnetometers. The core result is the analytic solution of the corresponding nonlinear magnetic field equation. As a precondition the current in the lines must be known.

The Aerial Core Project\footnote{https://aerial-core.eu/} is aimed at the maintenance and monitoring of large infrastructure. It includes, among other things, the maintenance of high-voltage lines using aerial manipulation. Examples of this are the setting of spacers for high-voltage lines and helical or clip-type bird diverters. Some of these actions require high forces to be generated by the manipulator, and for this purpose additional manipulator would be added to the system, where one manipulator would be used to hold the power line conductor while the other manipulator performs the required manipulation operation. These operations can generally be carried out while the power line is in operation, generating a strong alternating magnetic field around the conductor. This could cause interference in the electronics and sensors of the aerial manipulator. The idea of this paper is to show how it is possible to use this magnetic field as an alternative localization method of high-voltage line.

There are several systems that rely on the magnetic field for localization and navigation. Most of the work is designed for indoor environments with special infrastructure for magnetic navigation. Such an approach is shown in \cite{brown1973} and \cite{Everett1995} for vehicle navigation, which follows a wire that conducts alternating current laid under the floor based on two components of the magnetic field. This type of navigation is available in commercial autonomous guided vehicles. A similar approach is presented in \cite{Kamewaka1987}, where the magnetic guidance band is placed on the floor for localization, which is detected by a magnetic flux sensor.

In \cite{martinovic_electric_2014} a positioning system for inductive charging of electric vehicles was developed. For this purpose, at least two magnetometers are mounted on the underbody of the vehicle to sense the low-frequency magnetic field emitted by the charging coil in the parking lot. The vehicle then locates the charging coil by applying the trilateration principle to the measured data \cite{martinovic_magnetic_2015}. Since the magnetization effects in the ferromagnetic underbody can be compensated \cite{martinovic_electric_2014-1},\cite{martinovic_dealing_2019}, the assistance system achieves millimeter accuracy.

Several state of the art localization solutions rely on magnetic beacons distributed in the environment for 2D and 3D localization, such as the systems presented in \cite{Sheinker2013}, \cite{Sheinker2016}, \cite{Sheinker20132} and \cite{Mitterer2018}. A similar approach is presented in \cite{Son2016} for 5D localization, where the robot is equipped with magnets and the Hall effect sensors are distributed in the environment.
\begin{figure}[t]
	\centering
	\includegraphics[width=0.995\linewidth]{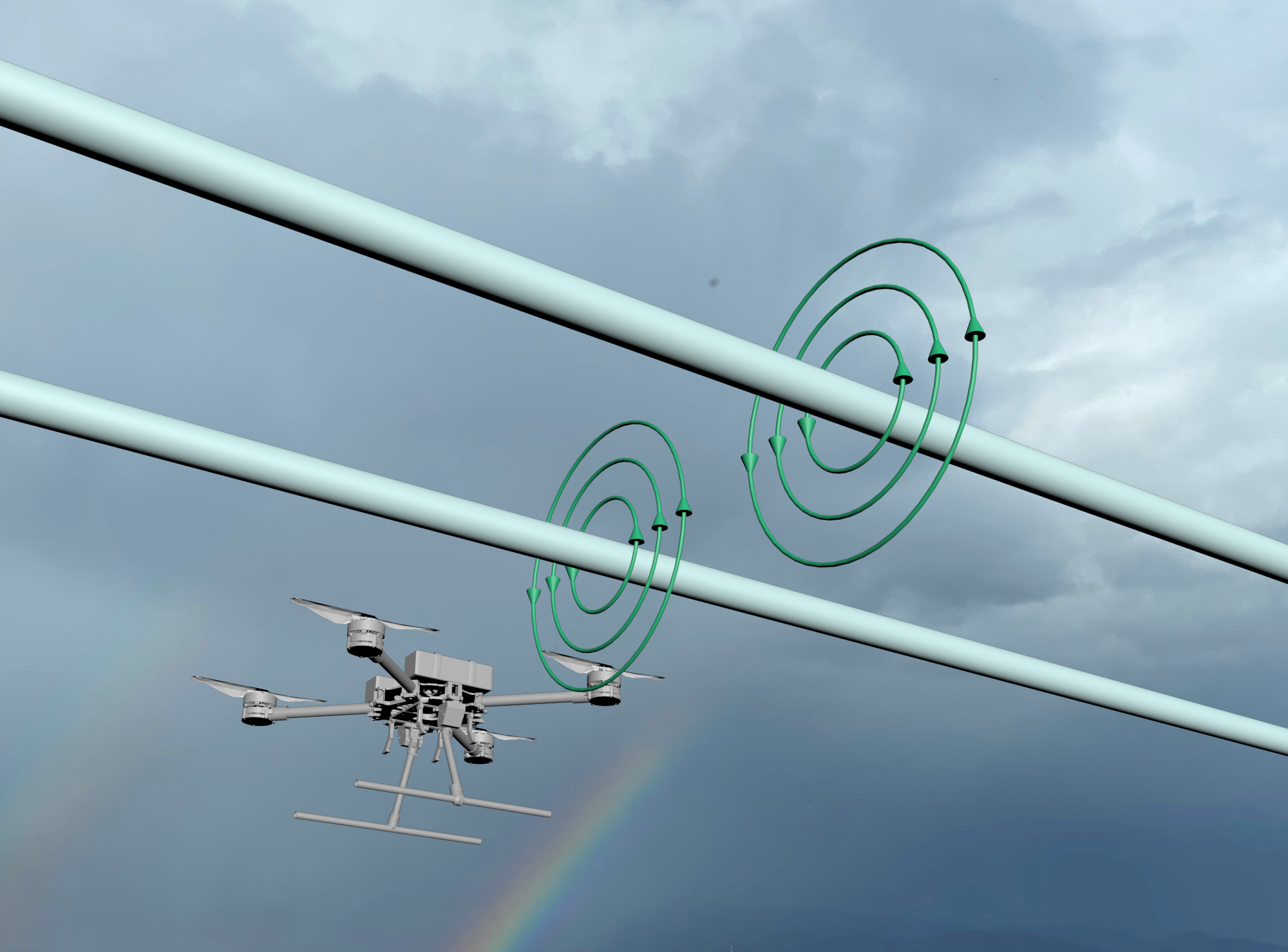}
	 \caption{UAV flying next to the operational power line}
     \label{fig:aerial_core}
\vspace{-0.5cm}
\end{figure}
A special type of localization based on magnetic fields is used in medicine to track robotic endoscopic capsules. Such applications are shown in \cite{Than2012}, \cite{Song2014}, \cite{Popek2017}, \cite{Hu2005}, \cite{hu2010} and \cite{Hu2007}. The number of works deals with localization based on the geomagnetic field maps of the environment. Examples of such work are \cite{Vallivaara2011}, \cite{Wang2016}, \cite{Lee2018}, \cite{Akai2017}, \cite{Frassl2013} and \cite{Hanley2017}.

In \cite{Wu2019}, the authors present the reconstruction of overhead line parameters for inspection by drones, with experiments performed in the laboratory without a drone. 

The real experiments with the drone touching 15-kV power lines are shown in \cite{Suarez2021}, together with the presentation of magnetometer readings.

In this paper, we present a validation of UAV localization of two-wire overhead lines based on magnetic field strength. We present two localization methods, an analytical one, which we have already presented in \cite{martinovic_mathematical_2021}, and a numerical one based on parameter optimization. The analytical method is fast and computationally cheap, but can only solve the problem with two wires, while the numerical method is much more computationally expensive, but can handle a different number of wires. The methods are tested on data collected from UAVs flown near AC power lines and validated based on the Optitrack system.

Section \ref{sec:localization} presents the localization methods used in the work. Section \ref{sec:setup} describes the experimental setup used for data collection, while section \ref{sec:results} shows the results. The conclusion is presented in \ref{sec:conclusion}.
\section{MAGNETIC LOCALIZATION}
\label{sec:localization}

We present the results of two localization methods based on magnetometer measurements, one analytical and one numerical. The analytical method is based on \cite{martinovic_mathematical_2021}, while the numerical method is based on the optimization of the criterion function that computes the magnetometer measurements based on the expected positions of the wires. The optimization parameters are the expected positions of the wires, while the error value is the difference between the actual measurements and the expected measurements.

\subsection{Analytical method for magnetic localization}
\label{subsec:analytical}
Assuming the model of very long, thin, and parallel transmission conductors, and the current they carry is of the same magnitude and direction, then according to \cite{martinovic_mathematical_2021} they produce a magnetic flux density distribution in space that can be calculated using the Biot-Savart law for thin conductors. Solving this law gives
\begin{equation}
	\mathbf{B}(y,z) =C \left(
	\begin{matrix}
		0\\
		\frac{-z}{z^2+(y+y_0)^2}+\frac{-z}{z^2+(y-y_0)^2}\\[6pt]
		\frac{y+y_0}{z^2+(y+y_0)^2}+\frac{y-y_0}{z^2+(y-y_0)^2}
	\end{matrix}\right), C=\frac{I\mu_0}{2\pi},
	\label{eq:field}
\end{equation}
where $C$ is a prefactor that depends on the transmission line current, $\mu_0$ is the magnetic field constant, and \mbox{$\pm y_0$} describes the position of the lines in the $xy$-plane. It follows that long lines have no field in the longitudinal or $x$ direction. This property can be exploited to determine the orientation of the UAV relative to the overhead line system. Assuming that the 3-axis magnetometers mounted on the UAV are aligned with their local frame $L_u$, the measurement $\mathbf{B_{mi}}$ of the $i$-th sensor in the global transmission line frame $\lbrace G \rbrace$ can be determined with
\begin{equation}
	\mathbf{B}(\mathbf{p_i})=\mathbf{R^{L_u}_G}(\alpha,\beta,\gamma)\,\mathbf{B_{mi}},
	\label{eq:connectVi2Bi}
\end{equation}
where $\mathbf{R^{L_u}_G}$ is the rotation matrix with $\alpha$, $\beta$ and $\gamma$ being the yaw, pitch, and roll angle of the UAV. The vector $\mathbf{p_i}$ is the position of the sensor $i$ in $\lbrace G \rbrace$. From \eqref{eq:field} it follows with the Cartesian unit vector $\mathbf{e_x}$ the equation for the $x$-component which is zero, i.\,e.:
\begin{equation}
0=(\mathbf{R^{L_u}_G},\mathbf{B_{mi}})\cdot \mathbf{e_x}
\end{equation}
If the rotation matrix is chosen to be \mbox{$\mathbf{R_G^{L_u}}(\alpha,\beta,\gamma)$=$\mathbf{R_x}(\gamma) \, \mathbf{R_y}(\beta) \, \mathbf{R_z}(\alpha)$}, the equation contains only the yaw and pitch angle which can be calculated analytically. Therefore two sensor measurements $\mathbf{B_{ml}}$ and $\mathbf{B_{mn}}$ of the sensors $l$ and $n$ are necessary. Solving the equation system leads to the two final equations for the yaw and pitch angle
\begin{equation}
	\alpha=atan\left(\frac{(\mathbf{B_{ml}}\times \mathbf{B_{mn}})\cdot \mathbf{e_y}}{(\mathbf{B_{ml}}\times \mathbf{B_{ml}})\cdot \mathbf{e_x}}\right)
	\label{eq:yawAng}
\end{equation}
and further with $A=tan\,\alpha$ to
\begin{equation}
	\beta=atan\left(\frac{B_{mnx}+B_{mny}A}{B_{mnz}\sqrt{A^2+1}}\right)=atan\left(\frac{B_{mlx}+B_{mly}A}{B_{mlz}\sqrt{A^2+1}}\right)
	\label{eq:pitchAng}
\end{equation}
Using the same approach to calculate the roll angle $\gamma$ ends up in a linear dependent equation system which does not provide any solution. The state of the arts currently is that it cannot be calculated analytically. However, during the assembly work typically the UAV's roll is very small so in the following it is assumed to be zero.

Now having all three angles either by above calculations and assumptions or by otherwise measurement, the position of the particular sensors in the $yz$-plane can be determined. Thereby the $x$-coordinate cannot be found as \eqref{eq:field} is independent of $x$. This is a natural consequence from the fact that there is no flux in $x$-direction. To get the sensor position the equation \eqref{eq:connectVi2Bi} is used. First, the right side is calculated which transforms the measurement of sensor $i$ into the global frame $\lbrace G \rbrace$. The resulting $\mathbf{B_i}$ then is connected to the left side via \eqref{eq:field} leading to the equation system
\begin{equation}
	\mathbf{B}(\mathbf{p_i})
	=\left(\begin{matrix}
		B_y(y_i,z_i)\\
		B_z(y_i,z_i)%
	\end{matrix}\right)
	=\left(\begin{matrix}
		B_{iy}\\
		B_{iz}%
	\end{matrix}\right)
	=\mathbf{B_i}
	\label{eq:finEqSysForPos}
\end{equation}
According to \cite{martinovic_mathematical_2021} the most simple way to solve \eqref{eq:finEqSysForPos} is described by the following steps. First, from the magnetic energy \mbox{$P_i^2$=$\vert \mathbf{B_i} \vert^2$=$\mathbf{B}(y_i,z_i)\cdot \mathbf{B}(y_i,z_i)$} the quadratic expression $z_i^2$ is determined. If omitting the sensor index $i$ for better readability it is
\begin{equation}
		z^2(y,P)=\pm 2\sqrt{y^2 y_0^2-\frac{C^2 y_0^2}{P^2}+\frac{C^4}{P^4}}-y_0^2-y^2+\frac{2C^2}{P^2}.%
	\label{eq:zsolSigPwr}
\end{equation}
Second, the expression is inserted into the $z$-component of \eqref{eq:finEqSysForPos}, i.\,e. \mbox{$B_z(y,z^2(y,P))$=$B_z$}. From this follows a fourth order polynomial for the $y$-component of the corresponding sensor which can be calculated analytically. The two real solutions are
\begin{equation}
		y_{1,2}=\frac{1}{P^2}\left(\pm\sqrt{r}\,cos \frac{\varphi}{2} +CB_z\right),
	\label{eq:yFinalSol}
\end{equation}where $r$ and $\varphi$ are
\begin{equation}
    \begin{matrix}
	    r=\sqrt{a^2+b^2},\ \ \varphi=tan^{-1}\left(\frac{b}{a}\right)+\pi,\\[6pt]
	    a=P^4 y_0^2+(2 B_z^2-P^2)C^2,\ b=2 C^2 B_z \sqrt{P^2-B_z^2}.
	 \end{matrix}
	\label{eq:r_and_phi}
\end{equation}
In the next step for each $y$-solution two $z$-solutions via \eqref{eq:zsolSigPwr} can be calculated, i.\,e. there are four possible locations in the $yz$-plane a sensor can be, which comes from the symmetry properties of the resulting magnetic field. It cannot be determined which one is the correct one. However, in \cite{martinovic_mathematical_2021} a method has been described which combines the solution of several sensors to multiple polygons in the $yz$-plane. All possible polygons in the last step then can be compared to the original polygon which is known by the user defined sensor positions in the local UAV frame. From this follow the sensor positions and the UAV position. Details can be read in \cite{martinovic_mathematical_2021} and are beyond the scope of this paper. In this paper, the choice of the correct solution is  based on the numerical criteria presented in the following subsection.

\subsection{Numerical method for locating two power lines based on magnetometer measurements}

All magnetometer measurements $\mathbf{B_{mi}}$ are measured in magnetometer frames $\mathbf{M_i}$ defined by the translation vector $\mathbf{p_i}$ and the rotation matrix $\mathbf{R^{L_u}_G}$ (assuming that all magnetometers have the same orientation) with respect to the global coordinate frame $G$.

From the coordinate frame of UAV, we can define two parallel conductors as two lines in 3D, that can be described by the following equations:
\begin{equation}
\begin{array}{ll}
    \mathbf{p_{pl}^1}=\mathbf{p_{pl0}^1}+\mathbf{v_{pl}}s\\
    \mathbf{p_{pl}^2}=\mathbf{p_{pl0}^2}+\mathbf{v_{pl}}s    
\end{array}
\label{eq:line}
\end{equation}

where $\mathbf{p_{pl}^1}=\begin{bmatrix} x_{pl}^1 &  y_{pl}^1 & z_{pl}^1\end{bmatrix}^T$ defines points of a first line , $\mathbf{p_{pl}^2}=\begin{bmatrix} x_{pl}^2 &  y_{pl}^2 & z_{pl}^2\end{bmatrix}^T$ defines points of a second line , $\mathbf{p_{pl0}^1}=\begin{bmatrix}x_{pl0}^1 & y_{pl0}^1 & z_{pl0}^1\end{bmatrix}^T$ is one point on the first line, $\mathbf{p_{pl0}^2}=\begin{bmatrix}x_{pl0}^2 & y_{pl0}^2 & z_{pl0}^2\end{bmatrix}^T$ is one point on the second line, $\mathbf{v_{pl}}=\begin{bmatrix}v_{xpl0} & v_{ypl0} & v_{zpl0}\end{bmatrix}^T$ is a line direction vector common for both lines and $s$ is scalar.

Assuming that the conductors are parallel to each other, their direction vector $\mathbf{v_{pl}}$ can be determined by the cross product of the vectors of the individual measurements. In the case of multiple magnetometers, this can be achieved by summing all combinations of the cross products of the individual magnetometers. It is important to note that in this way, the measurements that provide a better value of the direction vector $\mathbf{v_{pl}}$ have a greater influence.
For the case of 4 magnetometers, $\mathbf{v_{pl}}$ can be determined by the following equation:

\begin{equation}
\begin{array}{ll}
    \mathbf{v_{pl}}=\mathbf{B_{m1}}\times \mathbf{B_{m2}}+\mathbf{B_{m1}}\times \mathbf{B_{m3}}+\mathbf{B_{m1}}\times \mathbf{B_{m4}}+\\
    +\mathbf{B_{m2}}\times \mathbf{B_{m3}}+\mathbf{B_{m2}}\times \mathbf{B_{m4}}+\mathbf{B_{m3}}\times \mathbf{B_{m4}}
\end{array}
\label{eq:powerlinedirection}
\end{equation}

\begin{figure}[tb]
\centering
\includegraphics[width=0.85\columnwidth]{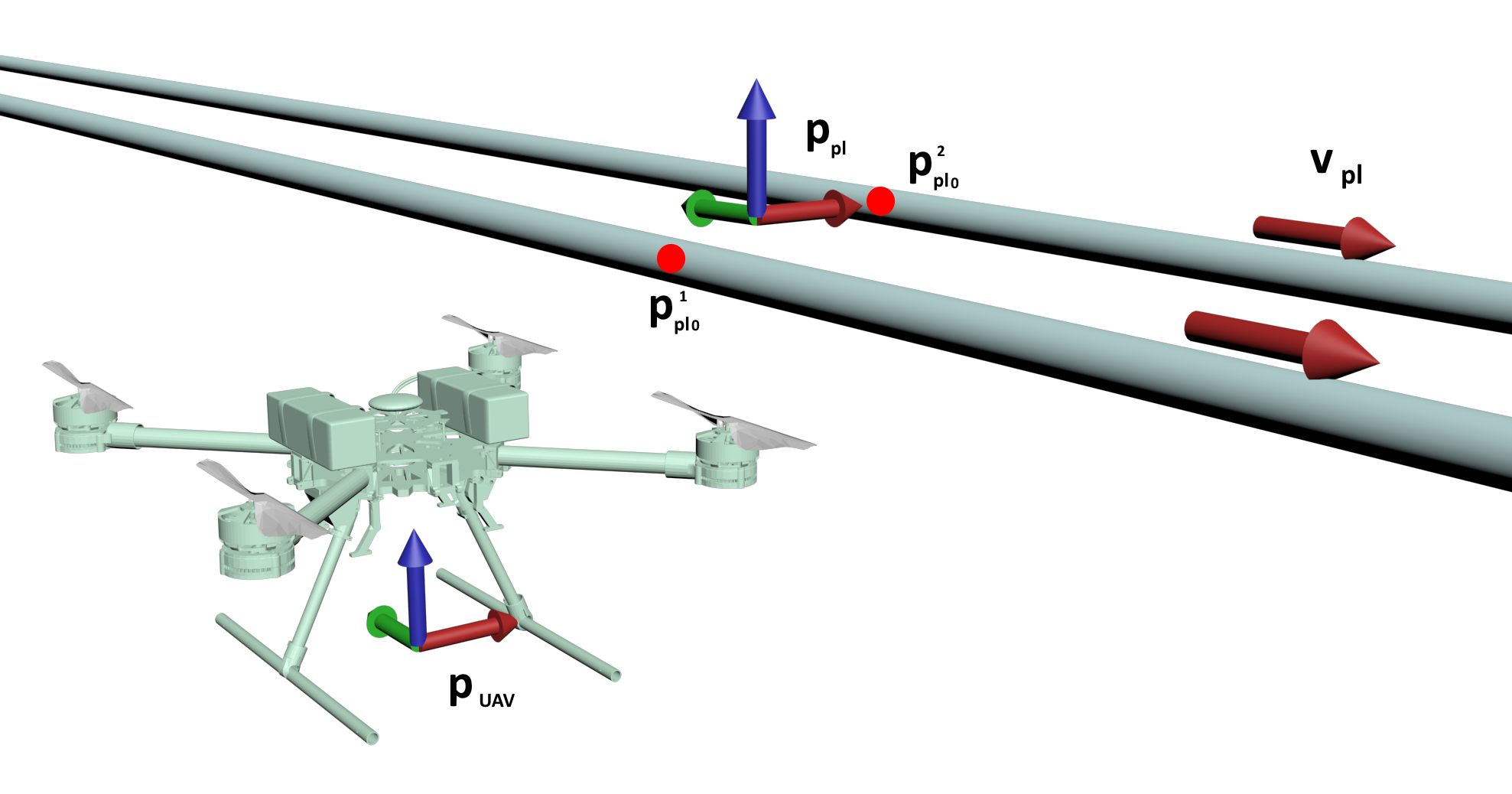}
\caption{Coordinate frames used for localization. The red arrow represents the $x$ direction, the green arrow represents the $y$ direction, and the blue arrow represents the $z$ direction.}
\label{fig:frames}
\end{figure}

With two conductors and 4 magnetometers, the numerical localization of the conductors can be done by minimizing the following criteria:

\begin{equation}
    J(\mathbf{p_{pl}^1},\mathbf{p_{pl}^2})=\sum_{i=1}^4|\mathbf{B_{mi}}-\mathbf{R^{L_u}_G}^{-1}\mathbf{B}(\mathbf{p_{mi}})|
\label{eq:criteria1}
\end{equation}

where $\mathbf{p_{mi}}$ is location of magnetometer $i$ in the UAV local coordinate frame (see figure \ref{fig:frames}).

Criteria \eqref{eq:criteria1} depend on 6 variables (3 in $\mathbf{p_{pl0}^1}$ and 3 in $\mathbf{p_{pl0}^2}$) which can be  reduced since we have assumed, that the conductors are parallel to each other and to the ground and at the known distance, as well as the fact that $\mathbf{p_{pl0}^1}$, $\mathbf{p_{pl0}^2}$ can be arbitrary points on conductors defined by \eqref{eq:line}. 

$\mathbf{p_{pl0}^1}$ can be defined by two independent variables, but first, the vector perpendicular to $\mathbf{v_{pl}}$ must be found:
\begin{equation}
    \mathbf{v_{plperp}}=\mathbf{v_{pl}} \times \begin{bmatrix}0 & 0 & 1\end{bmatrix}^T
\end{equation}
where $\mathbf{v_{plperp}}$ is a vector perpendicular to the $\mathbf{v_{pl}}$ and $z$ vectors. Now $\mathbf{p_{pl0}^1}$ can be represented with variables:
\begin{equation}
\mathbf{p_{pl0}^1}=s_1\begin{bmatrix}0 & 0 & 1\end{bmatrix}^T+s_2\mathbf{v_{plperp}}
\label{eq:xpl1}
\end{equation}
where $s_1$ and $s_2$ are two scalar variables. 

$\mathbf{p_{pl0}^2}$ can be determined by:
\begin{equation}
    \mathbf{p_{pl0}^2}=\mathbf{p_{pl0}^1}+d_c\mathbf{v_{plperp}}
\label{eq:xpl2}
\end{equation}
where $d_c$ is distance between power lines.

Given the equations \eqref{eq:xpl1} and \eqref{eq:xpl2}, the equation \eqref{eq:criteria1} becomes:

\begin{equation}
    J(s_1,s_2)=\sum_{i=1}^4|\mathbf{B_{mi}}-\mathbf{R^{L_u}_G}^{-1}\mathbf{B}
        (\mathbf{p_{mi}})|
\label{eq:criteria2}
\end{equation}

The problem boils down to the minimization of two variable criteria. In order to avoid jumps in the results, caused by the measurements noise, additional part is added to the criteria in order to  punish jumps from results in the previous step. The final criteria is:

\begin{equation}
\begin{array}{rr}
    J(s_1^t,s_2^t)=\sum_{i=1}^4|\mathbf{B_{mi}}-\mathbf{R^{L_u}_G}^{-1}\mathbf{B}
        (\mathbf{p_{mi}})|\cdot\\
        \cdot(1+q(|s_1^t-s_1^{t-1}|+|s_2^t-s_2^{t-1}|))
\end{array}
\label{eq:criteria3}
\end{equation}
where $s_1^t$ and  $s_2^t$ are scaling parameter $s_1$ and $s_2$ in moment $t$ and $q$ is the weight of the result in previous step.

This minimization can be done by various methods. We use Nelder-Mead simplex optimization from several different starting point to minimize the function \eqref{eq:criteria2}.

The criteria \eqref{eq:criteria3} is additionally used to select correct value out of several possible analytical solutions as presented in subsection \ref{subsec:analytical}.

\subsection{Processing the measurements of the alternating magnetic field}

When conductors carry alternating current, as in the case of power lines, the magnetic field produced is also alternating. In order to use magnetic field measurements, they must first be processed, in the sense that the amplitude ($A$), phase offset ($\phi$), and DC ($D$) components of the signal are determined. The alternating magnetic field can be described by:
\begin{equation}
    B_{alt}(t)=D+A\cos{(2\pi ft+\phi)}
\end{equation}
where $B_{alt}$ is the magnetic field , $f$ is the signal frequency (usually 50 or 60 Hz) and $t$ is the time.

In our previous work \cite{Vasiljevic2021}, we presented the method for extracting phase and amplitude based on the discrete Fourier transform, which has been shown to be effective for a large number of samples and processes information at 1 Hz. To increase the sampling rate of the system, we have developed an improved numerical optimization based processing system which is numerically more expensive but requires a smaller number of samples to determine $A$, $D$ and $\phi$. The optimization criteria used are:

\begin{equation}
    J_{alternating}(A,D,\phi)=\sum_{i=1}^N{(m_i-B_{alt}(i\delta_t))^2}
\end{equation}

where $J_{alternating}(A,D,\phi)$ is the optimization criterion, $m_i$ is the i-th of $N$ samples used for processing, and $\delta_t$ is the sampling time. The optimization is performed using the Nelder-Mead optimization algorithm, which is run twice for two different initial parameter sets $(A_0,D_0,\phi_0)$ for optimization.

The same optimization procedure is repeated for all 4 magnetometers used for each of the axes $x$, $y$ and $z$. If the phases of two axes match, they have the same direction, if their phases are apart by $\pi$ radians, they have opposite direction. For this, all measurements must be synchronized.
\section{EXPERIMENTAL SETUP}
\label{sec:setup}
An experimental setup is prepared to test magnetic field-based localization.

\subsection {Conductor setup}

\begin{figure}[t]
	\centering
	\includegraphics[width=0.95\linewidth]{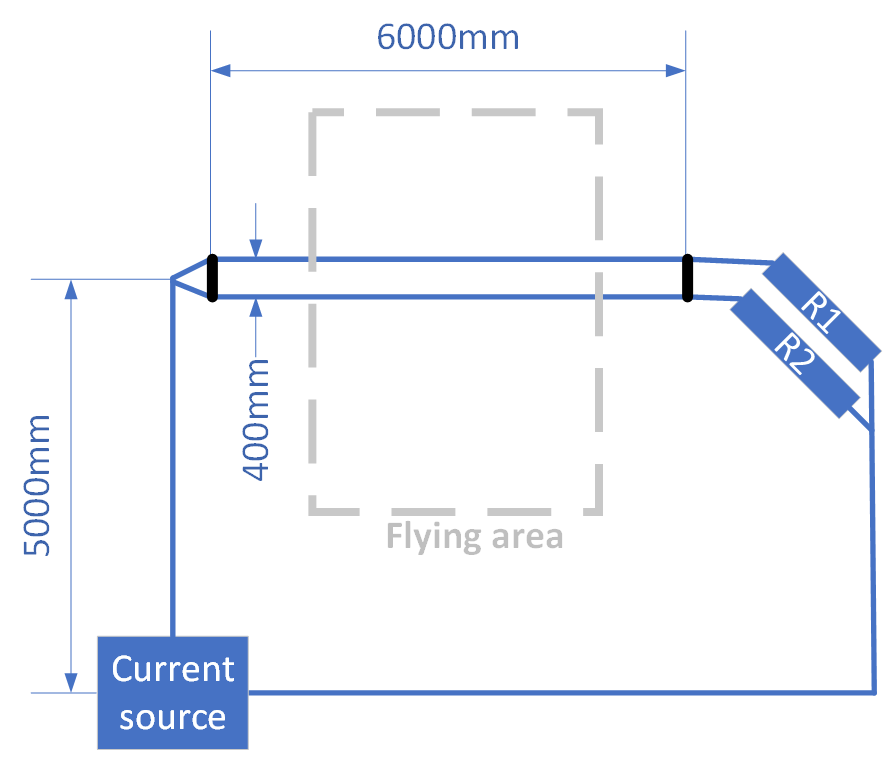}
	\includegraphics[width=0.95\linewidth]{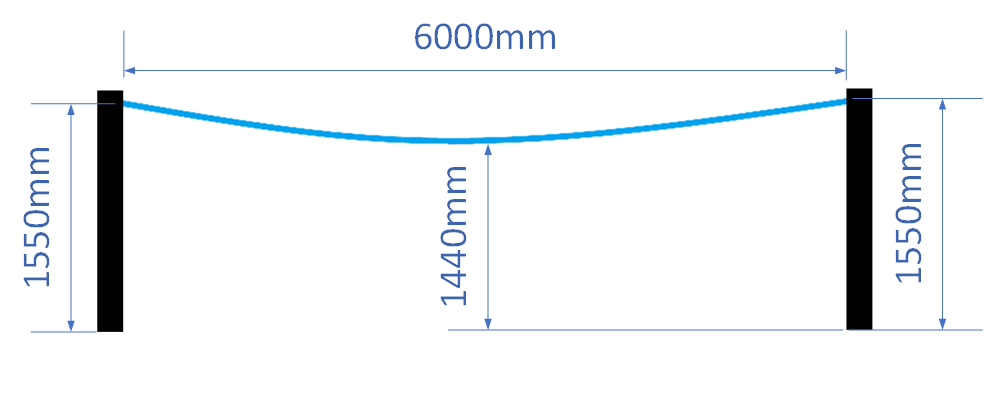}
	 \caption{Wires setup}
     \label{fig:setup1}
\vspace{-0.5cm}
\end{figure}

\begin{figure}[t]
	\centering
	\includegraphics[width=0.95\linewidth]{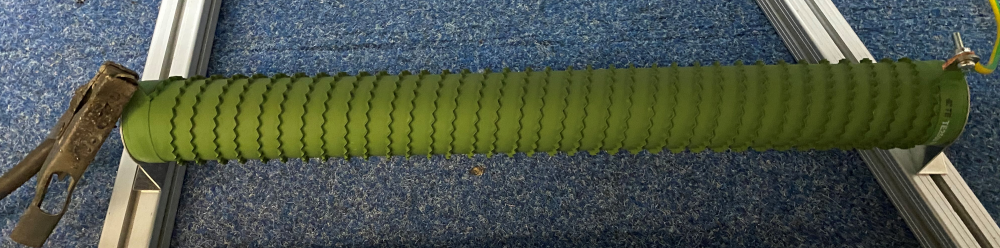}
	 \caption{2.5kW 1 ohm resistor}
     \label{fig:resistor}
\vspace{-0.5cm}
\end{figure}

\begin{figure}[t]
	\centering
	\includegraphics[width=0.8\linewidth]{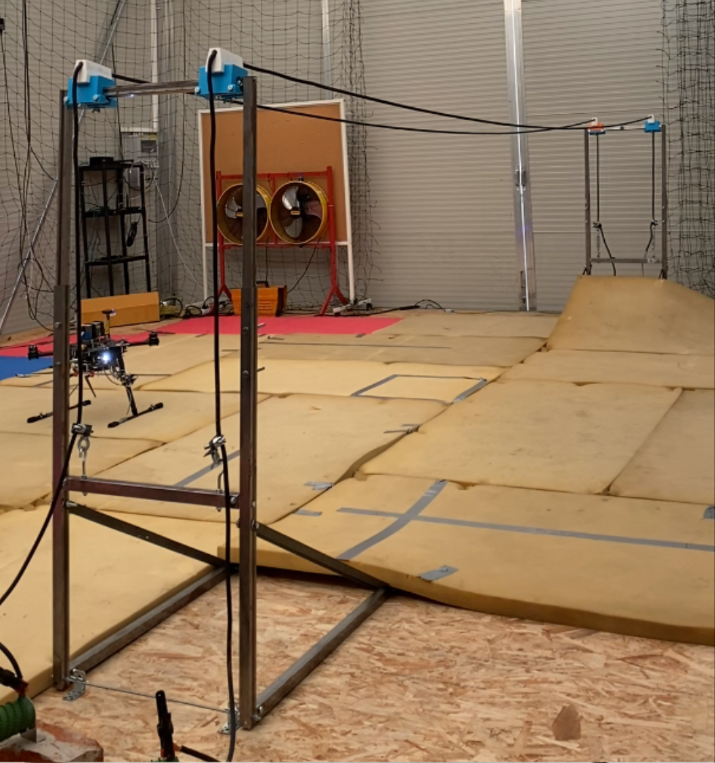}
	 \caption{Picture of conductor setup}
     \label{fig:setup_real}
\vspace{-0.5cm}
\end{figure}

The setup consists of two wires carrying alternating current at a frequency of 50 Hz. The power source is an electric welder consisting of a transformer that allows it to control the output current. The machine can output 140A rms current. The welder is the power source for both wires, in such a way that the output is divided into two wires, which are then laid in parallel from one part of the room to the other with a total length of 6 m with the help of a special stand at a distance of 40 cm (see Figure \ref{fig:setup1}). The wires are laid at a height of 1.55 m next to the stands and with a fall of 10 cm in the middle (see Figure \ref{fig:setup1}). The wires used are $25mm^2$ HO1N2-D insulated wires. After the stand, each wire is connected to 1 ohm resistors of 2.5 kW (see Figure \ref{fig:resistor}). After the resistors, the return wire is routed to the other side of the room and returned to the power source at a distance of 5 m from the two parallel wires. The current through each wire is 31A rms, measured with a current clamp on each wire. The picture of the conductor setup is shown in Figure \ref{fig:setup_real}.

\subsection{Optitrack setup}

Throughout the room, 13 Optitrack Prime 13 cameras are set to cover most of the room. Prior to the experiments, the cameras were calibrated to an accuracy of less than 1 mm. The sampling rate of the cameras was set to 250 Hz.

The Optitrack coordinate system is configured so that its $y$ axis coincides with the direction of the wire. In this way, only the $x$ and $z$ components of the measurements are relevant for comparison with the magnetometer-based localization system.

\subsection{Magnetometers}
\begin{figure}[t]
	\centering
	\includegraphics[width=0.95\linewidth]{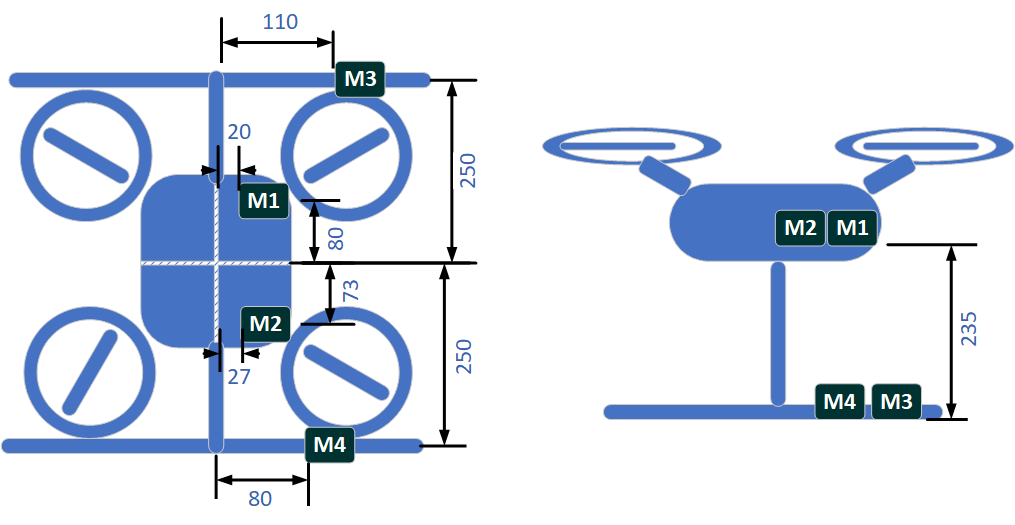}
	 \caption{Location of magnetometer sensors on UAV}
     \label{fig:uav}
\vspace{-0.5cm}
\end{figure}

For magnetic localization, we used four LIS3MDL 3-axis magnetometers that provide magnetic field strength measurements with a configurable range of ±4 Gauss to ±16 Gauss, which can be read out via a digital I$^2$C or SPI interface. Communication with the sensors is provided by the Arduino MKR Zero board via the SPI interface. All four sensors are connected to an Arduino and sensor selection is done via the chip select pin. In this way, a sampling rate of 500 Hz was achieved for all four sensors. Arduino collects the information from all four sensors and transmits it to the on-board computer via USB using a virtual serial port. For each sample from all four sensors, Arduino transmits four readings and a timestamp. The position of the magnetometers on the UAV is shown in Figure \ref{fig:uav}. Magnetometer positions are chosen to be as far away as possible from power cables and motors, but also to cover the largest polygon in the x-z plane \cite{martinovic_mathematical_2021}.

\subsection{UAV}

Used UAV is Kopterworks custom made quadrotor with 4 T-motor P60 KV170 motors. 
Autopilot used is Pixhawk CubeBlack, and the on board computer is Intel Nuc 10. During the experiments, the UAV was manually piloted around the wires and onboard computer was only used for data logging optitrack information received from the optitrack computer, based on the IR markers distributed around the UAV. The magnetometers were also distributed around the UAV in a way to achieve highest signal to noise ratio during pose estimation. Additionally, sensors had to be installed further away from motor supply cables to reduce noise caused by the motors.

All data collection is handled by on-board computer installed ROS. Two main nodes are used, one to receive location of the UAV from optitrack system and the second one which receives magnetometer information from Arduino. Information about flight telemetry were not used. During experiments topics are recorded using rosbag package.
\section{RESULTS}
\label{sec:results}

Using the setup presented in section \ref{sec:setup}, a flight test was performed.

\begin{figure}[t]
	\centering
	\includegraphics[width=0.95\linewidth]{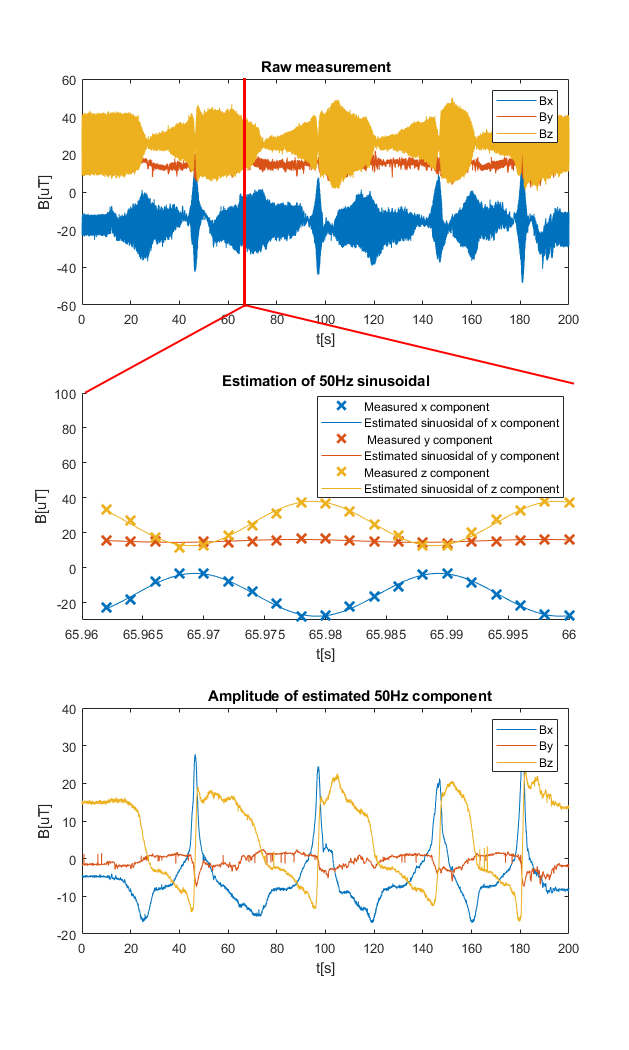}
	 \caption{Extraction of the 50 Hz component of the magnetic field strength from magnetometer 2. The top figure shows the raw measurements, the middle figure shows the estimate of the measurements with a 50 Hz sine function, and the bottom figure shows the estimated 50 Hz component}
     \label{fig:sinusoidal}
\end{figure}

Figure \ref{fig:sinusoidal} shows the raw magnetic field strength measurements and the extracted 50Hz components of the signal from magnetometer 2. The DC component of the signal in each direction represents the Earth's magnetic field strength.

\begin{figure}[t]
	\centering
	\includegraphics[width=0.95\linewidth]{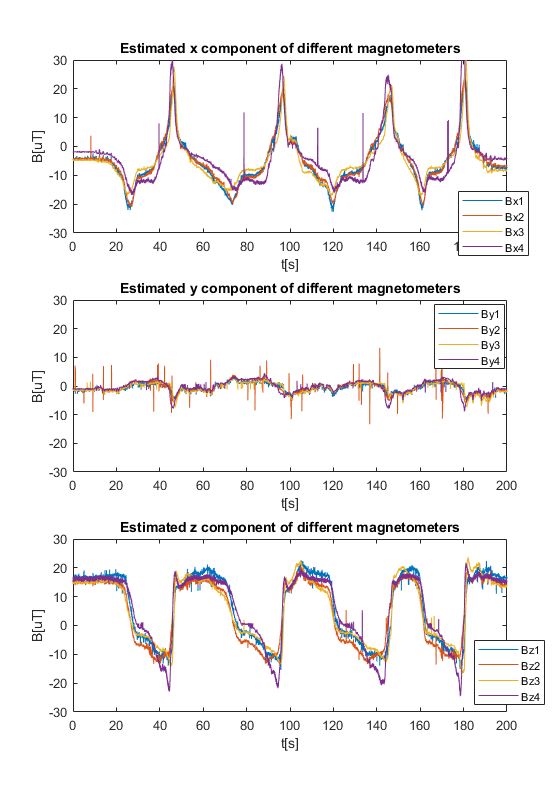}
	 \caption{Estimated x,y and z components of strength of 50Hz component of the magnetic field}
     \label{fig:magnetometers_components}
\end{figure}

All of the following calculations are based on the strength of the 50 Hz component of the magnetic field. Figure \ref{fig:magnetometers_components} shows the individual components of the 50 Hz magnetic vector estimated for different magnetometers. 

\begin{figure}[t]
	\centering
	\includegraphics[width=0.95\linewidth]{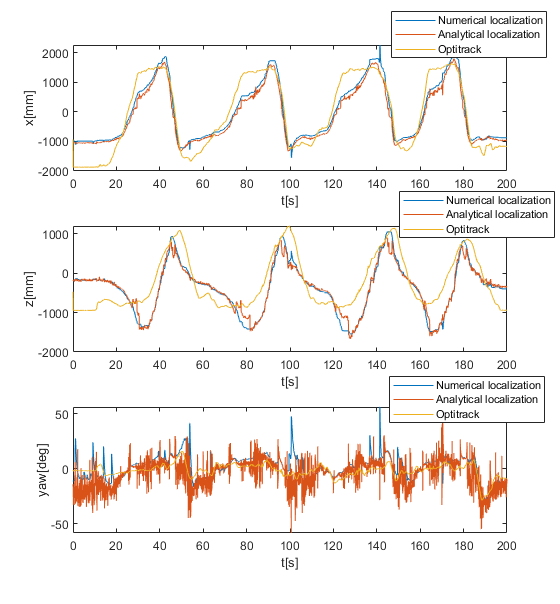}
	 \caption{2 wire estimate and measured individual $x$ and $z$ components and yaw angle of the UAV}
     \label{fig:localization_t}
\end{figure}

\begin{figure}[t]
	\centering
	\includegraphics[width=0.95\linewidth]{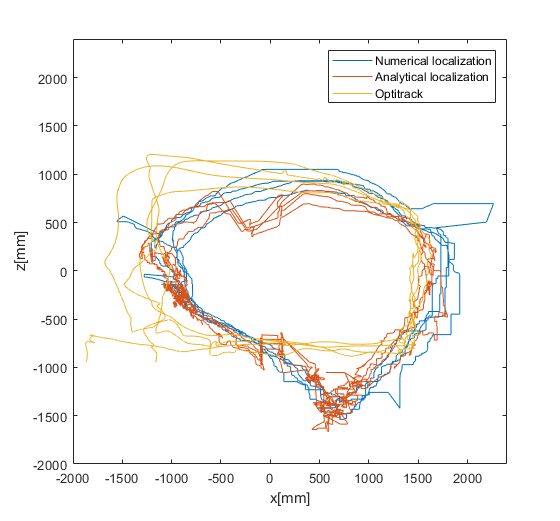}
	 \caption{2 wire estimate and measured $x$ - $z$ path of the UAV}
     \label{fig:localization_yz}
\end{figure}

Based on the magnetic field strength of 4 different magnetometers, the location of the UAV can be estimated. Based on the methods presented in section \ref{sec:localization}, for the case of two parallel wires 400 mm apart, the location of the UAV is calculated based on the magnetic field. Figure \ref{fig:localization_t} shows the individual $x$ and $z$ components of the position as well as the estimated yaw angle of the drone compared to the Optitrack measurements, while Figure \ref{fig:localization_yz} shows the position of the UAV in the $x$-$z$ coordinate system. 

The Matlab implementation of the analytical method data processing takes about 3 ms to localise in one time instance, while the numerical method processing takes about 350 ms.


\begin{figure}[t]
	\centering
	\includegraphics[width=0.95\linewidth]{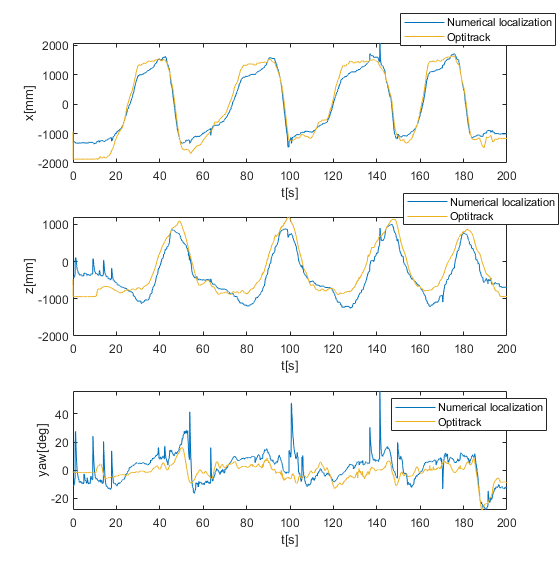}
	 \caption{3 wire estimate and measured individual $x$ and $z$ components and yaw angle of the UAV}
     \label{fig:localization_t_3}
\end{figure}

\begin{figure}[t]
	\centering
	\includegraphics[width=0.95\linewidth]{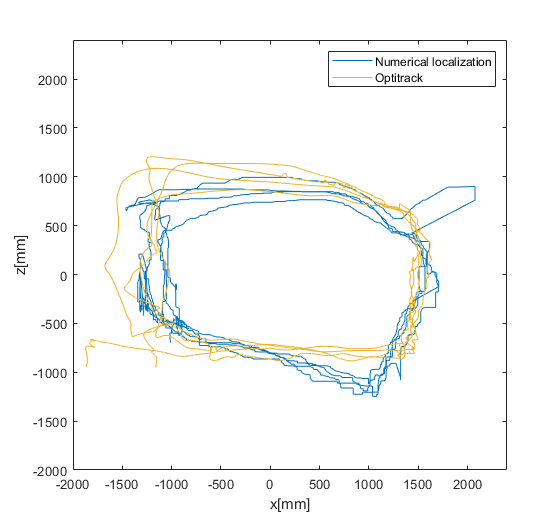}
	 \caption{3 wire estimate and measured $x$ - $z$ path of the UAV}
     \label{fig:localization_yz_3}
\end{figure}

 The same calculation for the numerical method is repeated for the case where 3 wires are considered, i.e., 2 original wires and a return wire parallel to them at a distance of 5000 mm from the first wire and with a $z$ coordinate lower by 1500 mm (see Figure \ref{fig:setup1}). The analytical method cannot be easily extended for this type of setup.
 The results are shown in Figures \ref{fig:localization_t_3} and \ref{fig:localization_yz_3}.
 
 The results have shown that magnetic field-based localization can provide stable location of the UAV  relative to the power line. Even if the absolute localization is not completely correct, the relative realation is preserved. The numerical and analytical methods have the same position profile, indicating that the additional influence of the ambient magnetic field disturbs the correct results. The numerical calculation with 3 wires has shown that the elimination of one of these influences (third wire) leads to a much better localization result.
 

\begin{figure}[t]
	\centering
	\includegraphics[width=0.8\linewidth]{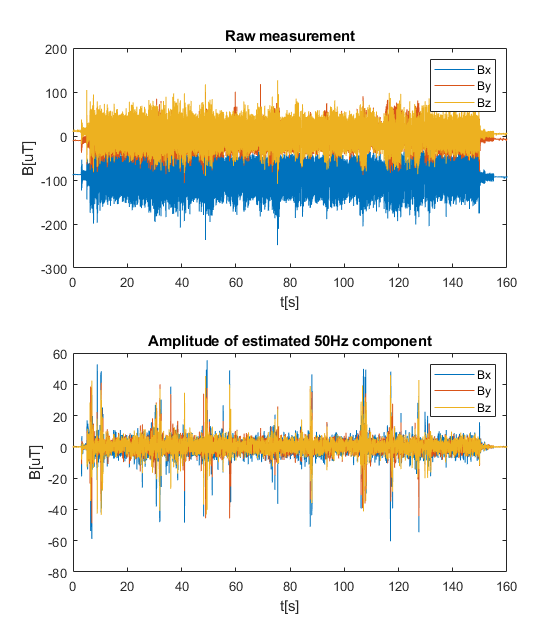}
	 \caption{Measurements and extracted 50 Hz component for magnetometer located next to the motor power cable during flight without external magnetic field}
     \label{fig:noise}
\end{figure}

An additional experiment was performed to show the influence of the noise of the motor power cable in the case of a poorly positioned magnetometer. In this experiment, the magnetometer was positioned right next to the motor cable. The magnetometer measurements during the flight without external magnetic field are shown in Figure \ref{fig:noise}. It can be seen that the noise from the motor drastically increases the noise of the 50 Hz component. For this reason, the magnetometers were placed as far away as possible from the motors and power cables, as shown in Figure \ref{fig:uav}.

\section{CONCLUSIONS}
\label{sec:conclusion}

In this paper, we presented validation of UAV localization of two-wire power lines using magnetic field strength. We present localization results based on two methods: the analytical method, which is based on explicit equations for UAV localization using magnetic field, and the numerical localization, which is based on optimization criteria that incorporate the locations of the wires. The methods were tested in a laboratory experiment in which a UAV equipped with four magnetometers flew around 2 wires carrying AC current. The results obtained were compared with ground truth collected by the Optitrack localization system. The results are promising and could enable autonomous flight based on magnetic field-based localization.

%

\section*{ACKNOWLEDGMENT}
This work was supported by the project  AERIAL COgnitive Integrated Multi-task Robotic System with Extended Operation Range and Safety (AERIAL CORE) EU-H2020-ICT (grant agreement No. 871479), and project Autonomous System for Assessment and Prediction of Infrastructure Integrity (ASAP), ESI ERDF - European Regional Development Fund

\bibliographystyle{IEEEtran}
\bibliography{bibliography.bib}

\begin{thebibliography}{10}
\providecommand{\url}[1]{#1}
\csname url@samestyle\endcsname
\providecommand{\newblock}{\relax}
\providecommand{\bibinfo}[2]{#2}
\providecommand{\BIBentrySTDinterwordspacing}{\spaceskip=0pt\relax}
\providecommand{\BIBentryALTinterwordstretchfactor}{4}
\providecommand{\BIBentryALTinterwordspacing}{\spaceskip=\fontdimen2\font plus
\BIBentryALTinterwordstretchfactor\fontdimen3\font minus
  \fontdimen4\font\relax}
\providecommand{\BIBforeignlanguage}[2]{{%
\expandafter\ifx\csname l@#1\endcsname\relax
\typeout{** WARNING: IEEEtran.bst: No hyphenation pattern has been}%
\typeout{** loaded for the language `#1'. Using the pattern for}%
\typeout{** the default language instead.}%
\else
\language=\csname l@#1\endcsname
\fi
#2}}
\providecommand{\BIBdecl}{\relax}
\BIBdecl

\bibitem{martinovic_mathematical_2021}
D.~Martinović, S.~Bogdan, and Z.~Kovačić,
  ``\BIBforeignlanguage{en}{Mathematical {Considerations} for {Unmanned}
  {Aerial} {Vehicle} {Navigation} in the {Magnetic} {Field} of {Two} {Parallel}
  {Transmission} {Lines}},'' \emph{\BIBforeignlanguage{en}{Applied Sciences}},
  vol.~11, no.~8, p. 3323, Jan. 2021, number: 8.

\bibitem{brown1973}
M.~Brown Boveri \& Cie~AG, ``Verfarhen und vorrichtung zur selbsta\"{a}tigen
  spurf\"{u}hrung von gleiselosen fahrzeugen,'' \emph{German pattent no:
  2137631}, 1973.

\bibitem{Everett1995}
H.~Everett, \emph{Sensors for mobile robot: Theory and application}.\hskip 1em
  plus 0.5em minus 0.4em\relax Wellesley, Massachusetts, NJ, USA: A K Peters,
  Ltd., 1995.

\bibitem{Kamewaka1987}
S.~Kamewaka and S.~Uemura, ``A magnetic guidance method for automated guided
  vehicles,'' \emph{IEEE Transactions on Magnetics}, vol.~23, no.~5, pp.
  2416--2418, 1987.

\bibitem{martinovic_electric_2014}
D.~Martinovic, M.~Grimm, and H.-C. Reuss,
  ``\BIBforeignlanguage{English}{Electric {Vehicle} {Positioning} {Concept} for
  {Inductive} {Charging} {Purposes} {Using} {Magnetic} {Fields}},'' in
  \emph{\BIBforeignlanguage{English}{{IEEE} {Power} \& {Energy} {Student}
  {Summit} 2014}}.\hskip 1em plus 0.5em minus 0.4em\relax Ostfildern: haka
  print und medien GmbH, 2014, pp. 11--16.

\bibitem{martinovic_magnetic_2015}
D.~Martinovic, C.~Binz, and H.-C. Reuss,
  ``\BIBforeignlanguage{English}{Magnetic {Field} based {Localization} of the
  {Charging} {Coil} using {Trilateration}},'' in
  \emph{\BIBforeignlanguage{English}{Autoreg 2015 - {VDI}-{Berichte}
  2233}}.\hskip 1em plus 0.5em minus 0.4em\relax Düsseldorf: VDI Verlag GmbH,
  2015, pp. 129--140.

\bibitem{martinovic_electric_2014-1}
D.~Martinovic, M.~Grimm, and H.-C. Reuss,
  ``\BIBforeignlanguage{English}{Electric {Vehicle} {Positioning} for
  {Inductive} {Charging} {Purposes} {Using} {Magnetic} {Field} {Distortion}
  {Elimination} in {High}-{Permeability} {Environments}},''
  \emph{\BIBforeignlanguage{English}{IEEE Transactions on Magnetics}}, vol.~50,
  no.~11, pp. 1--4, Nov. 2014.

\bibitem{martinovic_dealing_2019}
D.~Martinovic, ``Dealing with magnetization effects in {EV} positioning systems
  based on periodic magnetic signals,'' \emph{AIP Advances}, vol.~9, no.~3, p.
  035214, Mar. 2019, publisher: American Institute of Physics.

\bibitem{Sheinker2013}
A.~Sheinker, B.~Ginzburg, N.~Salomonski, L.~Frumkis, and B.~Z. Kaplan,
  ``{Localization in 2D using beacons of low frequency magnetic field},''
  \emph{IEEE Journal of Selected Topics in Applied Earth Observations and
  Remote Sensing}, vol.~6, no.~2, pp. 1020--1030, 2013.

\bibitem{Sheinker2016}
A.~Sheinker, B.~Ginzburg, N.~Salomonski, L.~Frumkis, B.~Z. Kaplan, and M.~B.
  Moldwin, ``{A method for indoor navigation based on magnetic beacons using
  smartphones and tablets},'' \emph{Measurement: Journal of the International
  Measurement Confederation}, vol.~81, pp. 197--209, 2016.

\bibitem{Sheinker20132}
A.~Sheinker, B.~Ginzburg, N.~Salomonski, L.~Frumkis, and B.~Z. Kaplan,
  ``{Localization in 3-D using beacons of low frequency magnetic field},''
  \emph{IEEE Transactions on Instrumentation and Measurement}, vol.~62, no.~12,
  pp. 3194--3201, 2013.

\bibitem{Mitterer2018}
T.~Mitterer, H.~Gietler, L.-M. Faller, and H.~Zangl, ``{Artificial Landmarks
  for Autonomous Vehicles Based on Magnetic Sensors},'' \emph{Proceedings},
  vol.~2, no.~13, p. 856, 2018.

\bibitem{Son2016}
D.~Son, S.~Yim, and M.~Sitti, ``{A 5-D Localization Method for a Magnetically
  Manipulated Untethered Robot Using a 2-D Array of Hall-Effect Sensors},''
  \emph{IEEE/ASME Transactions on Mechatronics}, vol.~21, no.~2, pp. 708--716,
  2016.

\bibitem{Than2012}
T.~D. Than, G.~Alici, H.~Zhou, and W.~Li, ``{A review of localization systems
  for robotic endoscopic capsules},'' \emph{IEEE Transactions on Biomedical
  Engineering}, vol.~59, no.~9, pp. 2387--2399, 2012.

\bibitem{Song2014}
S.~Song, B.~Li, W.~Qiao, C.~Hu, H.~Ren, H.~Yu, Q.~Zhang, M.~Q. Meng, and G.~Xu,
  ``{6-D magnetic localization and orientation method for an annular magnet
  based on a closed-form analytical model},'' \emph{IEEE Transactions on
  Magnetics}, vol.~50, no.~9, 2014.

\bibitem{Popek2017}
K.~M. Popek, T.~Schmid, and J.~J. Abbott, ``{Six-Degree-of-Freedom Localization
  of an Untethered Magnetic Capsule Using a Single Rotating Magnetic Dipole},''
  \emph{IEEE Robotics and Automation Letters}, vol.~2, no.~1, pp. 305--312,
  2017.

\bibitem{Hu2005}
C.~Hu, M.~Q. Meng, and M.~Mandal, ``{Efficient magnetic localization and
  orientation technique for capsule endoscopy},'' \emph{2005 IEEE/RSJ
  International Conference on Intelligent Robots and Systems, IROS}, pp.
  628--633, 2005.

\bibitem{hu2010}
C.~Hu, M.~Li, S.~Song, W.~Yang, R.~Zhang, and M.~Q. Meng, ``{A Cubic 3-Axis
  Magnetic Sensor Array for Wirelessly Tracking Magnet Position and
  Orientation},'' \emph{IEEE Sensors Journal}, vol.~10, no.~5, pp. 903--913,
  2010.

\bibitem{Hu2007}
C.~Hu, M.~Q. Meng, and M.~Mandal, ``{A linear algorithm for tracing magnet
  position and orientation by using three-axis magnetic sensors},'' \emph{IEEE
  Transactions on Magnetics}, vol.~43, no.~12, pp. 4096--4101, 2007.

\bibitem{Vallivaara2011}
I.~Vallivaara, J.~Haverinen, A.~Kemppainen, and J.~R{\"{o}}ning, ``{Magnetic
  field-based SLAM method for solving the localization problem in mobile robot
  floor-cleaning task},'' \emph{IEEE 15th International Conference on Advanced
  Robotics: New Boundaries for Robotics, ICAR 2011}, pp. 198--203, 2011.

\bibitem{Wang2016}
S.~Wang, H.~Wen, R.~Clark, and N.~Trigoni, ``{Keyframe based large-scale indoor
  localisation using geomagnetic field and motion pattern},'' \emph{IEEE
  International Conference on Intelligent Robots and Systems}, vol.
  2016-November, pp. 1910--1917, 2016.

\bibitem{Lee2018}
N.~Lee, S.~Ahn, and D.~Han, ``{AMID: Accurate magnetic indoor localization
  using deep learning},'' \emph{Sensors (Switzerland)}, vol.~18, no.~5, 2018.

\bibitem{Akai2017}
N.~Akai and K.~Ozaki, ``{3D magnetic field mapping in large-scale indoor
  environment using measurement robot and Gaussian processes},'' \emph{2017
  International Conference on Indoor Positioning and Indoor Navigation, IPIN
  2017}, vol. 2017-Janua, no. September, pp. 1--7, 2017.

\bibitem{Frassl2013}
M.~Frassl, M.~Angermann, M.~Lichtenstern, P.~Robertson, B.~J. Julian, and
  M.~Doniec, ``{Magnetic maps of indoor environments for precise localization
  of legged and non-legged locomotion},'' \emph{IEEE International Conference
  on Intelligent Robots and Systems}, no. May 2014, pp. 913--920, 2013.

\bibitem{Hanley2017}
D.~Hanley, A.~B. Faustino, S.~D. Zelman, D.~A. Degenhardt, and T.~Bretl,
  ``{MagPIE: A dataset for indoor positioning with magnetic anomalies},''
  \emph{2017 International Conference on Indoor Positioning and Indoor
  Navigation, IPIN 2017}, vol. 2017-Janua, pp. 1--8, 2017.

\bibitem{Wu2019}
Y.~Wu, G.~Zhao, J.~Hu, Y.~Ouyang, S.~X. Wang, J.~He, F.~Gao, and S.~Wang,
  ``{Overhead Transmission Line Parameter Reconstruction for UAV Inspection
  Based on Tunneling Magnetoresistive Sensors and Inverse Models},'' \emph{IEEE
  Transactions on Power Delivery}, vol.~34, no.~3, pp. 819--827, 2019.

\bibitem{Suarez2021}
A.~Suarez, R.~Salmoral, P.~J. Zarco-Perinan, and A.~Ollero, ``{Experimental
  Evaluation of Aerial Manipulation Robot in Contact with 15 kV Power Line:
  Shielded and Long Reach Configurations},'' \emph{IEEE Access}, vol.~9, pp.
  94\,573--94\,585, 2021.

\bibitem{Vasiljevic2021}
G.~Vasiljević, D.~Martinović, M.~Orsag, and S.~Bogdan, ``Grabbing power line
  conductors based on the measurements of the magnetic field strength,''
  \emph{2021 Aerial Robotic Systems Physically Interacting with the Environment
  (AIRPHARO)}, pp. 1--7, 2021.

\end{thebibliography}

\end{document}